\documentclass{article}

\usepackage[preprint]{neurips_2025}

\usepackage[utf8]{inputenc} 
\usepackage[T1]{fontenc}    
\usepackage{hyperref}       
\usepackage{url}            
\usepackage{booktabs}       
\usepackage{amsfonts}       
\usepackage{nicefrac}       
\usepackage{microtype}      
\usepackage{xcolor}         
\usepackage{latexsym}

\usepackage{inconsolata}
\usepackage{graphicx}
\usepackage{subfigure}
\usepackage{threeparttable}
\usepackage{adjustbox}
\usepackage{tcolorbox}
\usepackage{amsmath}
\usepackage{amssymb}
\usepackage{mathtools}
\usepackage{amsthm}
\usepackage{multirow}
\usepackage{bm}
\usepackage{tabularx}
\usepackage{booktabs}
\usepackage{caption}
\usepackage{adjustbox}
\tcbuselibrary{skins}
\usetikzlibrary{shadings}
\usepackage{array}
\usepackage{enumitem}
\usepackage{ulem}
\usepackage{wrapfig}
\usepackage{subcaption}
\usepackage{float}
\tcbuselibrary{breakable}

\theoremstyle{plain}
\newtheorem{theorem}{Theorem}[section]

\theoremstyle{definition}
\newtheorem{definition}[theorem]{Definition}

\theoremstyle{remark}

\title{Efficient Text-Attributed Graph Learning through Selective Annotation and Graph Alignment}

\author{
Huanyi Xie\textsuperscript{1,2},
Lijie Hu\textsuperscript{†,1,2},
Lu Yu\textsuperscript{3},
Tianhao Huang\textsuperscript{1}, \\
\textbf{Longfei Li\textsuperscript{3},
Meng Li\textsuperscript{3},
Jun Zhou\textsuperscript{3},
Huan Wang\textsuperscript{4},
Di Wang\textsuperscript{1,2}}\\
$^1$Provable Responsible AI and Data Analytics (PRADA) Lab\\
$^2$King Abdullah University of Science and Technology \\
$^3$Ant Group \quad 
$^4$Huazhong Agricultural University
}

\begin{document}

\maketitle

\begin{abstract}
In the realm of Text-attributed Graphs (TAGs), traditional graph neural networks (GNNs) often fall short due to the complex textual information associated with each node. Recent methods have improved node representations by leveraging large language models (LLMs) to enhance node text features, but these approaches typically require extensive annotations or fine-tuning across all nodes, which is both time-consuming and costly. To overcome these challenges, we introduce GAGA, an efficient framework for TAG representation learning. GAGA reduces annotation time and cost by focusing on annotating only representative nodes and edges. It constructs an annotation graph that captures the topological relationships among these annotations. Furthermore, GAGA employs a two-level alignment module to effectively integrate the annotation graph with the TAG, aligning their underlying structures. Experiments show that GAGA achieves classification accuracies on par with or surpassing state-of-the-art methods while requiring only 1\% of the data to be annotated, demonstrating its high efficiency.
\end{abstract}

\def\thefootnote{†}\footnotetext{Corresponding Author.}

\section{Introduction}
In real-world scenarios, many graphs can be effectively represented as text-attributed graphs (TAGs) \cite{yang2021graphformers}, where nodes correspond to textual data and edges signify relationships such as citations or purchases. TAGs are widely used in domains like text classification, recommendation systems, social networks, and fake news detection \cite{yang2015network, juan2023attentive,liu2025taad,zhang2024multi,xiang2024preserving,wang2023inductive,wang2025epm,hong2024multitask,wang2024persistent}. For instance, in the Ogbn-arxiv dataset, nodes represent academic papers, while edges denote citation links \cite{hu2020open}.

Traditional graph neural networks (GNNs), such as Graph Convolutional Network (GCN) \cite{kipf2016semi} and GraphSAGE \cite{hamilton2017inductive}, are commonly employed for tasks like node classification and link prediction on TAGs. However, these models often underutilize the rich textual information associated with nodes, which is crucial for context understanding, leading to suboptimal performance on large datasets like Ogbn-arxiv \cite{yan2023comprehensive}.

To address this, recent works leverage large language models (LLMs) to enhance textual features or refine graph structures \cite{he2023harnessing, sun2023large}. For example, TAPE \cite{he2023harnessing} generates node annotations via LLM-predicted summaries and explanations, while methods like SIMTEG \cite{duan2023simteg} fine-tune LLMs for node representations before training GNNs. Although effective, these approaches are often computationally expensive, requiring significant time and costs for annotation and fine-tuning. For instance, annotating the Ogbn-arxiv dataset with TAPE takes 9 hours and costs \$128.

To overcome these challenges, we propose GAGA: Graph Alignment Guided Text-attributed Graph Learning. Instead of annotating all nodes, GAGA identifies and annotates only representative nodes (or edges), significantly reducing time and costs. These annotations are used to construct an annotation graph, which is aligned with the TAG through a novel two-level contrastive learning process. This framework efficiently integrates semantic information from LLMs and topological information from GNNs.

GAGA comprises three stages: (1) selecting and annotating representative nodes or edges using LLMs to construct an annotation graph, (2) performing two-level contrastive learning between the annotation graph and the TAG, and (3) applying the learned representations to downstream tasks, where only the GNN is fine-tuned while the LLM remains frozen. Experimental results on six datasets demonstrate that GAGA achieves comparable or superior performance to state-of-the-art methods while reducing annotation requirements to just 1\% of the data, improving efficiency by $3\sim100$ times. Our contributions can be summarized as follows:
\begin{itemize}[noitemsep,topsep=1pt,parsep=3pt,partopsep=1pt]
    \item We propose GAGA, a lightweight framework for TAG representation learning. Compared to previous methods, GAGA considers an efficient strategy that only annotates representative nodes (edges), thereby reducing both the time and cost of annotation via LLMs. It recognizes and incorporates the topological relationships between these annotations by constructing an annotation graph. 
    
    \item Unlike previous methods, which are based on fine-tuning the whole dataset, GAGA leverages the alignment instead. To achieve this, we align sub-annotation graphs with the subgraphs in TAG. Specifically, we developed a two-level contrastive learning (sub-graph level and prototype level), which provides significant improvement in performance. These two components could also be used in other problems. 
    \item 
    We provide comprehensive experiments to demonstrate the performance of GAGA. Specifically, via experiments over six datasets, GAGA  achieves a classification accuracy that is comparable to
or even exceeds SOTA methods for node classification and link prediction tasks. Moreover, compared to previous LLM-augmented SOTA methods, GAGA only requires 1\% of the data to be annotated, making it surprisingly efficient. 
\end{itemize}

\section{Related Work}
\paragraph{\bf Language Models for Text-Attributed Graph.}
Traditional GNN approaches typically handle TAGs by converting textual attributes into features via shallow neural networks, such as bag-of-words, which limits the comprehensive understanding of textual semantics. Recent research has focused on embeddings based on language models (LMs) like BERT \cite{devlin2018bert} to address this issue. These fine-tuning pre-trained models can effectively generate deep embeddings to capture the rich semantic information within TAGs. There are two main architectures for empowering TAG tasks using LMs. The first is the cascaded architecture, where the textual information of nodes is independently encoded by LMs, and GNN models then aggregate the outputs to obtain the final embeddings, including TextGNN \cite{zhu2021textgnn}, GEAR \cite{zhou2019gear}, GIANT \cite{chien2021node}, GPT-GNN \cite{hu2020gpt}, and SimTeG \cite{duan2023simteg}. However, this approach separates text encoding from graph aggregation, which prevents a unified integration of the two processes. To address this limitation, the nested architecture has also been widely studied. This approach integrates text encoding and graph aggregation, performing these tasks iteratively to better unify both processes. For example, Graphormer \cite{yang2021graphformers}, GLEM \cite{zhao2022learning} and DRAGON \cite{yasunaga2022deep} follow this nested architecture.

LLMs such as ChatGPT \cite{brown2020language} have shown tremendous potential in various NLP tasks. However, how to apply LLMs to graph-structured data, such as TAGs, remains challenging \cite{chai2023graphllm,qin2023disentangled}. \cite{chen2024exploring} investigated the potential of LLMs in node classification tasks. Some works have already attempted to use LLMs for TAGs. \cite{he2023harnessing} proposed TAPE, which leverages LLMs to use the explanations generated during zero-shot classification as informative features for the graph. Another category of methods utilizes LLMs' strengths in text understanding to improve the graph's topology by enhancing the semantic understanding of node information. A representative example of this approach is the work by \cite{sun2023large}, where they utilize LLMs to generate pseudo-labels and compute the semantic similarity between node attributes to refine the graph's topology through edge addition/removal and weight adjustment. \cite{yu2023empower} propose ENG that enhances TAGs by using LLMs to generate additional nodes and the corresponding additional information. Additionally, \cite{pan2024distilling} employs a knowledge distillation approach to distill LLMs into graph models specifically for TAG learning. However, focusing solely on either nodes or edges has its limitations. Concentrating only on nodes (textual information) can result in the loss of the network's topological structure, while focusing exclusively on edges makes it difficult to fully capture the semantic information of the nodes themselves. Therefore, we need a structure that combines both nodes and edges to enhance the TAG representation, which is a strength of GAGA. Moreover, as mentioned above, all previous SOTA methods are time-consuming and costly, while our GAGA is highly lightweight and achieves almost the same performance. 

\paragraph{\bf Explanation-Guided Learning.}
With the advancement of AI, the importance of research in Explainable Artificial Intelligence (XAI) has become increasingly prominent~\cite{hu2024faithful,hu2025towards,hu2023seat,lai2023faithful,hu2024improving,hu2024towards,hu2024editable,gou2023fundamental,hu2024semi}. Although researchers have made significant progress in XAI techniques, leading to more attempts to generate explanations for Deep Neural Networks (DNNs), more profound issues, such as how to apply XAI techniques to improve the performance of DNN models, warrant further attention as research progresses \cite{ross2017right, 10.1145/3644073}. In the field of computer vision, there has been extensive research on using explanation supervision to guide model training \cite{das2017human, linsley2018learning, qiao2018exploring, mitsuhara2019embedding, zhang2019interpretable, gao2022res, zhang2023magi, saha2023saliency}. In NLP tasks, explanations are primarily generated through attention mechanisms or by utilizing auxiliary generative models \cite{bao2018deriving, strout2019human, zhong2019fine, choi2020less, ghaeini2019saliency, jain2020learning, stacey2022supervising}. Recently, several explanation supervision frameworks for GNNs have also emerged. For instance, GNES, proposed by \cite{gao2021gnes}, optimizes model explanations and predictions through weak supervision and regularization of the model's explanations. Compared to these papers, in GAGA, we consider the annotation given by LLMs as an explanation, which can further guide us for contrastive learning.

\section{Preliminaries}
\paragraph{Text-Attributed Graph.}
A Text-attributed Graph (TAG) can be formally represented as \( G = (V, A, \{s_n\}_{n \in V}) \), where \( V \) is a set of \( N \) nodes, \( A \in \mathbb{R}^{N \times N} \) is the adjacency matrix, \( s_n \in D^{L_n} \) is a sequential text associated with node \( n \in V \), with \( D \) as the dictionary of words or tokens, and \( L_n \) as the sequence length.

\noindent  \textbf{Large Language Models and Prompting.} 
Prompts can take various forms, such as a single sentence or longer paragraphs, and may include additional information or constraints to guide the model's behavior. Let $\mathcal{M}$ be an LLM that takes an input sequence $x = (x_{1}, x_{2}, \ldots, x_{q})$ and produces an output sequence $y = (y_{1}, y_{2}, \ldots, y_{m})$. The model is typically trained to optimize the conditional probability distribution $p(y|x)$, which assigns a probability to each possible output sequence $y$ given $x$. To incorporate a prompt $w$ with the input sequence $x$, we can concatenate them into a new sequence $\hat{x} = (w, x_{1}, x_{2}, \ldots, x_{q})$. The conditional probability distribution $p(\hat{y}|\hat{x})$ is then computed using $\hat{x}$. Formally, the probability of the output sequence $y$ given $\hat{x}$ is:
\[
p(\hat{y}|\hat{x}) = \prod_{i=1}^{m} p(y_{i}|y_{<i}, \hat{x}),
\]
where $y_{<i}$ represents the prefix of the sequence $y$ up to position $i-1$, and $p(y_{i}|y_{<i}, \hat{x})$ denotes the probability of generating $y_{i}$ given $y_{<i}$ and $\hat{x}$.

\paragraph{GNNs for Node Classification and Link Prediction.} GNNs are utilized for both node classification and link prediction via leveraging the structural and feature information of graphs. In node classification, the objective is to assign labels to nodes based on their attributes and connections. This is achieved by updating each node's representation through a message-passing process defined by:
\[
h_i^k = f^k \left( h_i^{k-1}, \text{AGG} \left( \{ h_j^{k-1} : j \in \mathcal{N}_i \} \right) \right) \in \mathbb{R}^d,
\]
where \( h_i^k \) is the representation of node \( i \) at layer \( k \), \( \mathcal{N}_i \) represents the set of neighbors, and \( f^k \) is a neural network layer that integrates the previous layer's representation with aggregated information from neighbors via a function like sum.

In link prediction, the task is to predict the probability of an edge existing between two nodes, \( i \) and \( j \), based on their learned representations and the overall graph structure. This probability is modeled by 
$p(i, j) = f(h_i^k, h_j^k),$
where the function \( f \) uses the representations derived from the node classification process to assess the likelihood between node \(i\) and \(j\).

\section{GAGA: Graph Alignment Guided Text-attributed Graph Learning}
\begin{figure*}[t]
  \centering
  \includegraphics[width=0.9\textwidth]{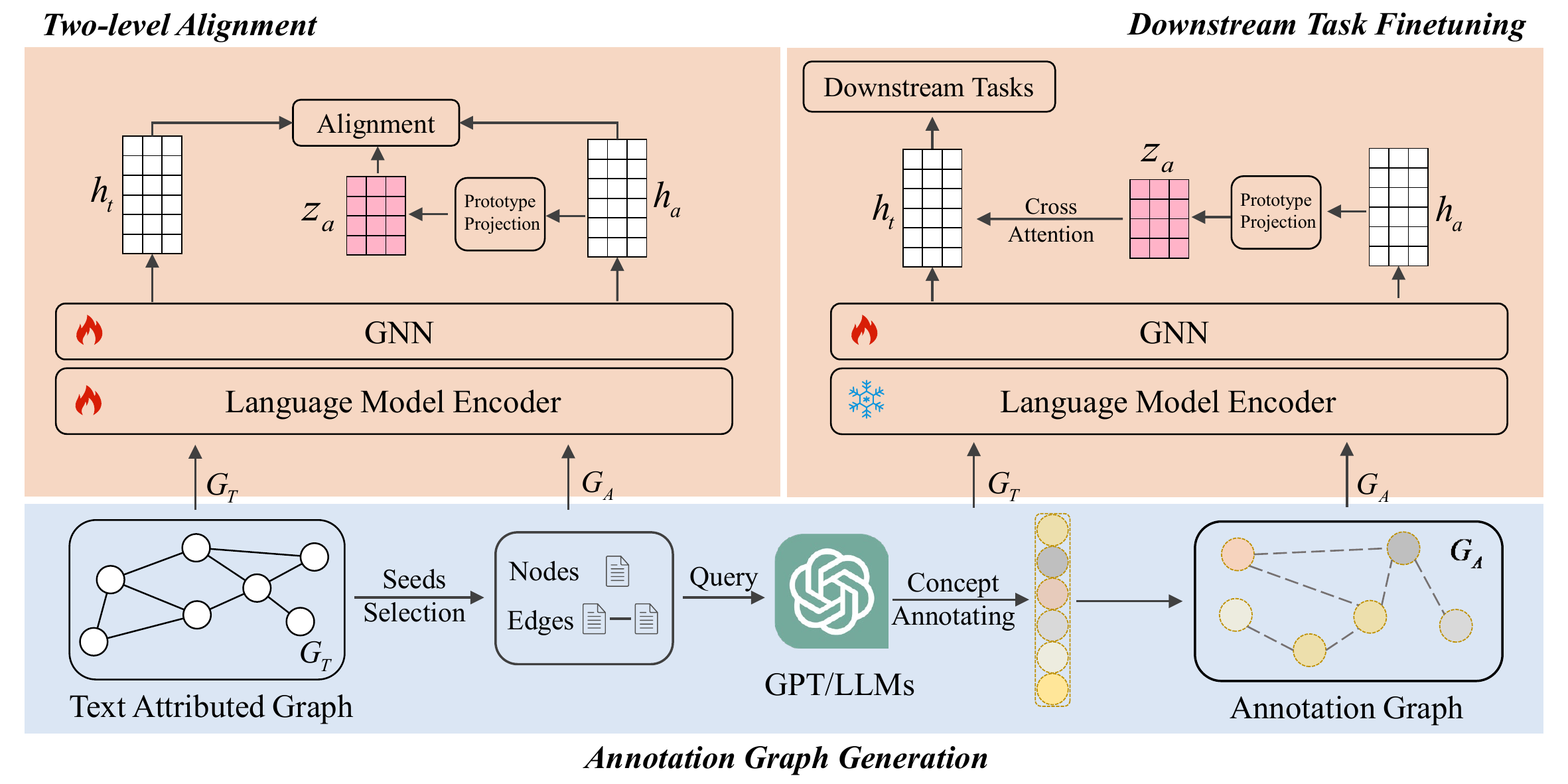}
  \vspace{-10pt}
  \caption{An overview of our GAGA framework. \label{fig:framework}}
  \vspace{-15pt}
\end{figure*}

In this section, we introduce our novel framework, GAGA. As illustrated in Figure \ref{fig:framework}, it primarily comprises three stages: annotation graph generation, two-level (subgraph) alignment, and downstream task fine-tuning. In the first stage, we first obtain a small but important set of nodes or edges through representative-based selection. Then, we generate annotations for these nodes or edges by prompting LLMs. These annotations are constructed into an annotation graph. In the second stage, we employ a two-level (subgraph level and prototype level) contrastive learning between the sub-annotation graphs and their corresponding sub-textual graphs, enabling lightweight fine-tuning of both the language model (LM) and GNNs using information from the selected nodes.  In the third stage, which involves downstream tasks, we only fine-tune the GNN while keeping the LM frozen. 

\subsection{Annotation Graph Generation}
The core idea of this stage is to use the information density metric to identify representative nodes and edges and then employ carefully designed prompts to utilize LLMs for annotation, thereby reducing costs.

\noindent  \textbf{Representative Node (Edge) Selection.}
As we mentioned earlier, using LLMs to annotate data can enhance model performance on TAGs. However, given that a typical graph contains tens of thousands of nodes, annotating each one can be both time-consuming and costly \citep{chen2023label}. Thus, to maintain high performance while annotating only a small number of nodes, it is crucial to select nodes (edges) with high information density \citep{cai2017active}. Information density, denoted as $\phi_{\text{density}}$, serves as a representative metric to identify nodes (edges) that best ``represent'' the underlying data distribution in the embedding space. To locate nodes situated in dense regions of the embedding space, we first perform $k$-means clustering on the embeddings of all unlabeled nodes. We then compute each node's Euclidean distance  to its nearest cluster center. The density score for a node $v_i$ is determined by converting this distance into a similarity score:
\begin{equation*}
    \phi_{\text{density}}(v_i) = \frac{1}{1 + \|\text{Emb}(v_i)-\text{Ce}(v_i)\| },
\end{equation*}
where $\text{Emb}(v_i)$ represents the embedding of node $v_i$, and $\text{Ce}(v_i)$ denotes the center of the cluster to which node $v_i$ belongs. Intuitively, a higher value of $\phi_{\text{density}}(v_i)$ indicates that node $v_i$ is closer to its center, making it more representative in the embedding space. We denote the selected subset of nodes as $V_a^* \subset V$, which consists of the nodes with top-$s$ highest information density, where $s$ is a hyperparameter. 

For link prediction tasks, we need to select a set of edges. For edge \(e_{i,j}\), we define its information density as the sum of the information densities of the nodes at both endpoints:
$\phi_\text{density}(e_{i,j}) = \phi_\text{density}(v_i) + \phi_\text{density}(v_j)$. Similar to nodes, the selection of the subset of edges, $V_a^*$, is based on the top-$\hat{k}$ highest information density. 

\noindent  \textbf{Annotation Generation.} 
Next, we will annotate these selected nodes (edges). In GAGA,  we prompt LLMs not only to produce predictions and explanations for the selected nodes but also to infer and understand the broader concepts and knowledge needed for these predictions and explanations. For edges, we prompt LLMs to explain the rationale behind their formation and to grasp the concepts and knowledge necessary to understand these connections. Our prompt templates can be found in Appendix \ref{App:prompts}. 

\noindent  \textbf{Annotation Graph Construction.}
To structurally represent the most informative knowledge contained in each node's text and leverage potential relationships between annotations, we construct an annotation graph $G^*_\mathcal{A} = (V^*_a, E^*_a)$. In detail, we first define each annotation for nodes or edges as a node $v_a \in V^*_a$, and then we build the annotation graph based on the similarity between each pair of nodes. For annotation node $v^i_a, v^j_a \in V_a^*$, the similarity $s(\cdot)$ is computed as follows:
\begin{equation*}
s(v^i_a, v^j_a) = \frac{ \text{Emb}(v^i_a) \cdot \text{Emb}(v^j_a)}{\|\text{Emb}(v^i_a)\|\cdot \|\text{Emb}(v^j_a)\|}. 
\end{equation*} 
We then apply the $k$-nearest neighbors (KNN) algorithm to capture the relationships between annotations, connecting each node to its $k'$ nearest neighbors with edges to form the annotation graph. Formally, the edges of the annotation graph $G^*_\mathcal{A}$ is defined as:
\begin{equation*}
E^*_a = \bigcup_{i} \{(v^i_a, v^j_a) \mid v^j_a \in N_{k'}(v^i_a)\}, 
\end{equation*} 
where $N_{k'}(v^i_a)$ is set of $k'$-nearest neighbors of annotation node $v^i_a$. 

\subsection{Two-level Alignment} 
As we have only annotated a subset of nodes (edges), how to generalize these annotations becomes the key question, which is the motivation of our two-level alignment. Our alignment consists of two levels: one is for aligning sub-textual graphs and sub-annotation graphs, while the other one is for aligning sub-textual graphs and the prototypes induced by sub-annotation graphs. 

\noindent  \textbf{Subgraph Alignment.}
To adapt the Language Model encoder for graph tasks, we perform graph alignment. Given that only a subset of nodes (edges) are annotated, we have to sample and align subgraphs based on this subset during this alignment process for better generalization. To achieve this, we use a straightforward approach: for each selected node $v^*$, we sample its $k$-hop neighbors in both the TAG and annotation graph $G^*_\mathcal{A}$ to form a sub-text graph $G_T$ and sub-annotation graph $G_A$, respectively. In link prediction, for selected edges $e^*$, the difference lies in including the $t$-hop neighbors of both nodes connected by the selected edges to form the sub-annotation graph.

Through subsampling, we can obtain pairs of \( G_T \) and \( G_A \) corresponding to each selected node (or edge). Our alignment goal is to maximize the similarity between each pair of \( G_T \) and \( G_A \), while simultaneously minimizing the similarity between non-corresponding pairs. Specifically, for each graph, nodes obtain their representations through a Language Model Encoder (LMEncoder), which are then integrated with graph structural information via a GNN. Mathematically, we have the following loss:  
\begin{equation}\label{eq:1}
 \mathcal{L}_1=  \sum_{i} (\| h_t^i - h_a^i \|^2 - \frac{1}{n-1} \sum_{j \neq i} \| h_t^i - h_a^j \|^2 ), 
\end{equation}
\[
h_T = \text{GNN}(X_T), h_A = \text{GNN}(X_A), 
\]
\[
X_T = \text{LMEncoder}(G_T), X_A = \text{LMEncoder}(G_A), 
\]
where \( h_T = \{ h_t^1, h_t^2, \ldots, h_t^n \} \) and  \( h_A = \{ h_a^1, h_a^2, \ldots, h_a^n \} \) is the embedding matrix for $G_T$ and $G_A$ with number of nodes $n$ in these subgraphs. 

\noindent \textbf{Prototype Alignment.}  As the number of selected nodes increases, the embedding matrix \( h_A \) of the annotation graph becomes increasingly large, resulting in higher computational complexity for downstream tasks. Additionally, the high-dimensional embeddings often contain significant redundancy, which can hinder the model's generalization ability. To tackle these issues,  we introduce prototypes via vector quantization (VQ)~\citep{van2017neural} to map annotation graph embeddings into a more compact prototype space. This not only reduces computational overhead but also enhances semantic clustering. 

Specifically, a prototype matrix \( Z_a \) is represented as a \( k_p \times d \) matrix, where \( k_p \) is significantly smaller than the original number of nodes \( n \). By fixing the number of prototypes, we effectively reduce the computational load of alignment. Specifically, each embedding vector \( h_a \) is mapped to its nearest prototype \( z_a \) in \( Z_a \):  
\begin{equation}\label{eq:proto}
    z_a = Z_a^m, \quad \text{where } \quad m = \arg\min_j \left( \| h_a - Z^j_a \|_2 \right),
\end{equation}  
where \( Z_a^m \) is the \( m \)-th row of \( Z_a \). This process enforces semantic aggregation, reducing redundancy while encouraging the model to learn more abstract and generalizable representations of the annotation graph. To better illustrate the necessity of prototypes, see Appendix \ref{sec:anno} for some visualization results. 
 
Building on these prototype projections, we construct a contrastive loss to align textual embeddings \( h_t \) with prototype embeddings \( z_a \). This loss is combined with the original contrastive loss into the final objective:  
\begin{align}\label{eq:3}
\mathcal{L} = & \alpha \sum_{i} \left[ \| h_t^i - z_a^i \|^2 - \frac{1}{n-1} \sum_{j \neq i} \| h_t^i - z_a^j \|^2 \right] \notag \\
&+ (1 - \alpha) \mathcal{L}_1. 
\end{align}
Here the hyperparameter \( \alpha \) governs the trade-off between textual and prototype alignment. A small \( \alpha \) may cause the model to overfit to redundant information in the original embeddings, while a large \( \alpha \) might lead to excessive reliance on prototypes, resulting in the loss of fine-grained semantic details. A balanced choice of \( \alpha \) allows the model to effectively leverage the compactness and generalization benefits of prototypes while retaining critical semantic information from the original embeddings. By aligning representations in both the specific concept space and the aggregated prototype space, our method achieves an optimal balance between computational efficiency and representational granularity.

\subsection{Downstream Task Finetuning}
For a given downstream task, we first use the frozen Language Model encoder to obtain the representations of nodes $X_T^{'}$, which are then processed through GNN to acquire the node embeddings, i.e., $h=\text{GNN}(X_T^{'})$. Finally, by computing cross attention with the prototype embeddings of the annotations, we integrate the information of the annotations and obtain the final node representation. 
\begin{equation*}
h' = \text{softmax}(\frac{hW^Q \cdot (Z_aW^K)^T}{\sqrt{d_k}}) \cdot Z_aW^V,  
\end{equation*}
where $W^Q$, $W^K$, $W^V$ are query, key, $Z_a$ is the matrix of prototypes in (\ref{eq:proto}), and value weight matrices and $d_k$ is the normalization factor. 

For the node classification task, we add a fully connected layer and then pass the output through a softmax function to obtain the probability distribution for each category.
\[
\hat{y} = \text{softmax}(f(h')), 
\] 
where \(f\) is a linear function. For the link prediction task, we use the element-wise multiplication of the representations of nodes \(a\) and \(b\) as the corresponding edge representation. This feature is then passed through a fully connected layer followed by a sigmoid function to obtain the probability of the edge's existence:
\[\hat{y} = \text{sigmoid}((h_a' \odot h_b') \cdot W).\]

\section{Experiments}
\subsection{Experimental Setting} \label{sec_exp_settings}
\noindent {\bf Datasets.} We choose six datasets for evaluation: Cora \citep{mccallum2000automating}, PubMed \citep{sen2008collective}, ogbn-arxiv \citep{hu2020open}, ogbn-products \citep{hu2020open,he2023harnessing}, and tape-arxiv23 \citep{he2023harnessing} for the node classification tasks and use Cora, Citeseer \citep{giles1998citeseer}, and PubMed for the link prediction tasks.
Details about the datasets can be found in Appendix \ref{App: datasets}. The detailed prompts used for generating concepts for each dataset are shown in Appendix \ref{App:prompts}. We split the datasets for training, validation, and testing following \citep{he2023harnessing} for node classification tasks and HeaRT \citep{li2023evaluating,mao2023revisiting} for link prediction tasks.

\noindent {\bf Baselines.}
In this study, we employed a variety of models for the node classification task to comprehensively evaluate the performance of different approaches on TAGs. Specifically, we selected the following 11 models:
(i) MLP;
(ii) {GCN}~\citep{kipf2016semi};
(iii) {GraphSAGE}~\citep{sun2021scalable};
(iv) {RevGAT}~\citep{li2021training};
(v) {InstructGLM}~\citep{ye2023natural};
(vi) {Graphormers}~\citep{ying2021transformers}; and
(vii) {TAPE}~\citep{he2023harnessing}
(viii) {GLEM}~\citep{zhao2022learning}
(ix) {SimTEG}~\citep{duan2023simteg}
(x) {ENGINE}~\citep{zhu2024efficient}
(xi) {GIANT}~\citep{chien2021node}. Further details about the baselines are given in Appendix~\ref{App:baselines}.

\begin{table*}[ht]
\centering
\caption{Comparison of Time (Minute) and Money Usage (\$) Highlighted by Efficiency and Accuracy. OOM refers to instances where memory usage exceeded 32G of GPU memory or 64G of system memory. Timeout refers to tasks that remained incomplete after 72 hours. We highlight the best results in \colorbox{green!30}{green}. \label{tab:usage_comparison} }
\resizebox{0.8\linewidth}{!}{
\begin{tabular}{@{}lccccccccccc@{}}
\toprule
& \multicolumn{3}{c}{\textbf{ogbn-arxiv}} & \multicolumn{3}{c}{\textbf{PubMed}} \\ 
\cmidrule(r){2-4} \cmidrule(r){5-7}
& \textbf{Time Cost $\downarrow$} & \textbf{Money Cost $\downarrow$} & \textbf{Accuracy $\uparrow$} & \textbf{Time Cost $\downarrow$} & \textbf{Money Cost $\downarrow$} & \textbf{Accuracy $\uparrow$} \\ 
\midrule
TAPE & 694 & 113.39 & 75.20 & 71 & 17.63 & 94.31 \\
GLEM & 446 & 7.81 & 76.04 & 50 & 0.87 & 92.57 \\
SimTEG & 1439 & 25.18 & 75.13 & 283 & 4.95 & 81.23 \\
OneForAll$^\ast$ & 185 & 3.24 & 0.6983 & 85 & 1.48 & 73.01 \\
ENGINE & OOM & OOM & OOM & 203 & 3.55 & 74.74 \\
InstructGLM & OOM & OOM & OOM & OOM & OOM & OOM \\
Graphormer  & 224 & 3.92 & 66.67 & 45 & 0.783 & 83.65 \\
GIANT & Timeout & Timeout & Timeout & 431 & 7.54 & 76.89 \\
GAGA(Ours) & \colorbox{green!30}{80} & \colorbox{green!30}{2.87} & \colorbox{green!30}{76.23} & \colorbox{green!30}{17} & \colorbox{green!30}{0.49} & \colorbox{green!30}{94.61} \\
\bottomrule
\end{tabular}
}
\end{table*}

\begin{table*}[ht]
\centering
\caption{Node classification performance comparison on different datasets (all values in \%). The best results are highlighted with \colorbox{blue!20}{\textbf{bold}}.}
\resizebox{0.8\linewidth}{!}{
\label{Exp: node-cls}
\begin{tabular}{lccccc}
\toprule
\textbf{Model} & \textbf{ogbn-arxiv} & \textbf{PubMed} & \textbf{Cora} & \textbf{ogbn-products} & \textbf{tape-arxiv23}\\
\midrule
MLP & 53.36 $\pm$ 0.15 & 86.35 $\pm$ 0.20 & 63.88 $\pm$ 0.12 & 53.85 $\pm$ 0.18 & 62.02 $\pm$ 0.25 \\
GCN & 71.82 $\pm$ 0.20 & 80.31 $\pm$ 0.15 & 88.24 $\pm$ 0.10 & 70.52 $\pm$ 0.12 & 63.41 $\pm$ 0.20 \\
SAGE & 71.71 $\pm$ 0.18 & 88.81 $\pm$ 0.15 & 89.11 $\pm$ 0.12 & 69.13 $\pm$ 0.20 & 64.30 $\pm$ 0.18 \\
RevGAT & 70.83 $\pm$ 0.15 & 88.50 $\pm$ 0.20 & 89.11 $\pm$ 0.10 & 69.64 $\pm$ 0.15 & 65.63 $\pm$ 0.20 \\
\midrule
Graphormer & 72.81 $\pm$ 0.18 & 88.24 $\pm$ 0.15 & 80.41 $\pm$ 0.20 & 68.15 $\pm$ 0.12 & 62.87 $\pm$ 0.25 \\
InstructGLM & 75.70 $\pm$ 0.20 & 93.84 $\pm$ 0.15 & 87.08 $\pm$ 0.12 & 65.32 $\pm$ 0.02 & 70.32 $\pm$ 0.12\\
GLEM & 75.60 $\pm$ 0.15 & 92.57 $\pm$ 0.20 & 74.69 $\pm$ 0.18 & 73.77 $\pm$ 0.15 & 78.58 $\pm$ 0.20 \\
SimTEG & 75.29 $\pm$ 0.18 & 81.20 $\pm$ 0.01 & 88.04 $\pm$ 0.12 & 74.51 $\pm$ 0.20 & 79.51 $\pm$ 0.15 \\
ENGINE & 76.02 $\pm$ 0.20 & 74.72 $\pm$ 0.15 & \colorbox{blue!20}{\textbf{91.48 $\pm$ 0.10}} & \colorbox{blue!20}{\textbf{80.05 $\pm$ 0.15}} &79.76 $\pm$ 0.20 \\
GIANT & 74.26 $\pm$ 0.15 & 76.90 $\pm$ 0.20 & 85.52 $\pm$ 0.18 & 74.06 $\pm$ 0.12 & 72.18 $\pm$ 0.25 \\
TAPE & 75.20 $\pm$ 0.18 & 94.31 $\pm$ 0.15 & 91.19 $\pm$ 0.12 & 79.96 $\pm$ 0.20 & 80.80 $\pm$ 0.18 \\
\midrule
GCN$^\ast$ & 62.17$\pm$1.03 & 79.83$\pm$0.29 & 61.19$\pm$4.11 & 37.90$\pm$0.59 & 51.52$\pm$0.18 \\
TAPE$^\ast$ & 69.02$\pm$0.16 & 91.78$\pm$0.12 & 83.76$\pm$2.27 & 35.53$\pm$0.21 & 51.99$\pm$0.62 \\
\midrule
GAGA(Ours) & \colorbox{blue!20}{\textbf{76.21 $\pm$ 0.15}} & \colorbox{blue!20}{\textbf{94.62 $\pm$ 0.20}} & 89.67 $\pm$ 0.12 & 78.87 $\pm$ 0.18 & \colorbox{blue!20}{\textbf{81.03 $\pm$ 0.25}}\\
\bottomrule
\end{tabular}}
\end{table*}

For link prediction tasks, we selected 4 types of models, comprising a total of 19 models, as baselines: heuristic models, embedding models, GNN models, and GNN+Pairwise Info models. Heuristic models use graph structure-based scores to predict links. Embedding models learn low-dimensional representations of nodes to estimate link likelihoods, GNN models leverage multi-hop graph structures through message passing, and GNN+Pairwise Info models enhance GNNs with additional node-pair-specific information to improve link prediction accuracy. More details are given in Appendix \ref{App:baselines}. 

\noindent {\bf Evaluation Metrics.} For node classification tasks, we use classification accuracy, time cost (min), and money cost of the whole framework as evaluation metrics. In the experiment of time and money comparison, time includes new textual feature annotation (like TAPE and GAGA), training and inference time. Money usage only includes the money spent on annotating new features with LLMs. For link prediction tasks, we choose MRR@10 and AUC as metrics. See details in Appendix \ref{App: metrics}.  

\noindent {\bf Experimental Setup.} 
In our experiments, we prompt GPT-3.5-turbo-1106 \citep{brown2020language} to generate annotations. The Language Model Encoder utilized is all-MiniLM-L6-v2 \citep{wang2020minilm}, and the GNN employed is a 4-layer GCN. During the annotation step, we consider $40$ clusters. For the node classification task, we use 1\% nodes for annotation and $\sqrt{n_\text{{edges}}}$ for link prediction annotation with edge number $n_\text{{edges}}$. During the alignment process, we set \(\alpha\) to 0.6, and prototype dimension $k_p$ to 40. We also use $2$-hop subgraphs for the alignment. We use the Adam optimizer with the learning rate $5e^{-5}$ for alignment and $1e^{-3}$ for downstream task fine-tuning. All experiments were conducted on a 32G V100 card with 10 CPU cores and a maximum memory of 64G (In Table \ref{Exp: node-cls}, we use 4xA100 GPUs to reproduce the results for ENGINE and InstructGLM, as they cannot be executed on a V100.).  We run each experiment for 5 times.

\subsection{Utility Analysis}
\noindent {\bf Time and Cost Comparison.} Table \ref{tab:usage_comparison} compares the time and cost (see Appendix~\ref{app:time_money_settings}) efficiency of GAGA with previous LLM-based methods on the ogbn-arxiv and PubMed datasets. GAGA proves highly efficient, requiring annotations for only 1\% of nodes, while other methods often annotate or fine-tune all nodes. TAPE incurs high costs due to API usage and extensive annotations. ENGINE, despite combining LLM and GNN, struggles with memory issues, as the Llama-3-7B model consumes substantial GPU resources, making it unsuitable for large datasets on a single GPU. GIANT uses self-supervised learning for node representations, but its pre-training is extremely time-intensive as node counts grow. GLEM and GraphFormers are relatively efficient among prior methods but are more costly than GAGA and fall short in accuracy compared to SOTA results and GAGA.

\noindent  \textbf{Node Classification.} 
In the node classification task, as shown in Table \ref{Exp: node-cls}, GAGA significantly outperforms previous classical GNN models, such as GCN, and transformer-based methods, such as Graphormer, across all datasets. Moreover, compared to LLM-based methods, our method still performs well; it achieves the best results on ogbn-arxiv, PubMed, and tape-arxiv23. While its performance on the Cora and ogbn-products datasets was second only to ENGINE and TAPE, the difference to the best, ENGINE, is less than 1.8\%. These results demonstrate that GAGA achieves a classification accuracy that is comparable to or even exceeds SOTA methods due to the high generalization ability of our alignment method. Additionally, when comparing the performance of TAPE and GCN using only 1\% of labeled data (GCN$^\ast$, TAPE$^\ast$), our two-level alignment method significantly outperforms them across all datasets.

\begin{table}[htbp]
\centering
\begin{minipage}{0.45\textwidth}
  \centering
  \caption{Accuracy (\%) with different backbones on ogbn-arxiv.}
  \label{Tab:backbones}
  \resizebox{0.7\linewidth}{!}{
  \begin{tabular}{lcc}
  \toprule
  \textbf{GNN} & \textbf{Test} & \textbf{Valid} \\
  \midrule
  MLP & 72.83 $\pm$ 0.32& 73.15 $\pm$ 0.12\\
  GCN & 76.21 $\pm$ 0.13& 77.25 $\pm$ 0.09\\
  SAGE & 75.93$\pm$ 0.02 & 76.11 $\pm$ 0.03\\
  GAT & 76.46 $\pm$ 0.22& 77.01 $\pm$ 0.34\\
  RevGAT & 75.66 $\pm$ 0.17& 75.87 $\pm$ 0.01\\
  \bottomrule
  \end{tabular}}
\end{minipage}\hfill
\begin{minipage}{0.45\textwidth}
  \centering
  \caption{Accuracy (\%) with different ratio (\%) of seed nodes on ogbn-arxiv.}
  \label{Tab:ratio}
  \resizebox{0.7\linewidth}{!}{
  \begin{tabular}{lcc}
  \toprule
  \textbf{Ratio} & \textbf{Test} & \textbf{Valid} \\
  \midrule
  0.1\% & 75.71 $\pm$ 0.02 & 77.01 $\pm$ 0.01\\
  0.2\% & 75.72 $\pm$ 0.01 & 77.00 $\pm$ 0.03\\
  0.4\% & 75.71 $\pm$ 0.01& 76.99 $\pm$ 0.00\\
  0.6\% & 75.71 $\pm$ 0.01& 76.99 $\pm$ 0.01\\
  0.8\% & 75.72 $\pm$ 0.02& 77.00 $\pm$ 0.01\\
  1.0\% & 76.21 $\pm$ 0.03& 77.25 $\pm$ 0.02\\
  \bottomrule
  \end{tabular}}
\end{minipage}
\end{table}

\textbf{Link Prediction.} Similar to Node Classification, we conducted experimental evaluations on the link prediction task, with the results presented in Table \ref{tab:linkpred}. Leveraging the annotations on selected edges and the two-level alignment, GAGA achieves near state-of-the-art performance across all evaluated datasets for both metrics, falling slightly short of NBFNet on the PubMed dataset for MRR@10.

When compared against different categories of models, GAGA consistently surpasses heuristic baselines (e.g., CN, Katz) by a substantial margin, highlighting the advantages of learned representations. It also generally outperforms embedding-based methods such as Node2Vec and MF, and standard GNN architectures like GCN, GAT, and SAGE. The most salient comparisons are with models incorporating GNNs and pairwise information. Here, GAGA demonstrates a competitive edge. For instance, while NCNC shows strong AUC on Cora, GAGA leads in MRR@10. Similarly, GAGA's comprehensive performance on Citeseer and its leading AUC on PubMed distinguish it within this advanced category. The results suggest that GAGA's approach to integrating edge annotations and two-level alignment provides a distinct advantage in capturing complex relational patterns essential for accurate link prediction. 

\begin{table*}
\vspace{-5pt}
\centering
\caption{Link prediction results on Cora, Citeseer, and PubMed. The best results are highlighted with \colorbox{blue!20}{\textbf{bold}}.}
\label{tab:linkpred}
\resizebox{0.98\linewidth}{!}{
\begin{tabular}{cccccccc}
\toprule
& \multirow{2}{*}{\textbf{Models}} & \multicolumn{2}{c}{\textbf{Cora}} & \multicolumn{2}{c}{\textbf{Citeseer}} & \multicolumn{2}{c}{\textbf{PubMed}} \\
\cmidrule{3-8}
& & \textbf{MRR@10} & \textbf{AUC} & \textbf{MRR@10} & \textbf{AUC} & \textbf{MRR@10} & \textbf{AUC} \\
\midrule
\multirow{5}{*}{\textbf{Heuristic}} 
& CN & 20.99 & 70.85 & 28.34 & 67.49 & 14.02 & 63.9 \\
& AA & 31.87 & 70.97 & 29.37 & 67.49 & 16.66 & 63.9 \\
& RA & 30.79 & 70.96 & 27.61 & 67.48 & 15.63 & 63.9 \\
& Shortest Path & 12.45 & 81.08 & 31.82 & 75.5 & 7.15 & 74.64 \\
& Katz & 27.4 & 81.17 & 38.16 & 75.37 & 21.44 & 74.86 \\
\midrule

\multirow{4}{*}{\textbf{Embedding}} 
& Node2Vec & 37.29 $\pm$ 8.82 & 90.97 $\pm$ 0.64 & 44.33 $\pm$ 8.99 & 94.46 $\pm$ 0.59 & 34.61 $\pm$ 2.48 & 93.14 $\pm$ 0.18 \\
& MF & 14.29 $\pm$ 5.79 & 80.29 $\pm$ 2.26 & 24.80 $\pm$ 4.71 & 75.92 $\pm$ 3.25 & 19.29 $\pm$ 6.29 & 93.06 $\pm$ 0.43 \\
& MLP & 31.21 $\pm$ 7.90 & 95.32 $\pm$ 0.37 & 43.53 $\pm$ 7.26 & 94.45 $\pm$ 0.32 & 16.52 $\pm$ 4.14 & 98.34 $\pm$ 0.10 \\
\midrule

\multirow{5}{*}{\textbf{GNN}} 
& GCN & 32.50 $\pm$ 6.87 & 95.01 $\pm$ 0.32 & 50.01 $\pm$ 6.04 & 95.89 $\pm$ 0.26 & 19.94 $\pm$ 4.24 & 89.69 $\pm$ 0.06 \\
& GAT & 31.86 $\pm$ 6.08 & 93.69 $\pm$ 0.27 & 48.69 $\pm$ 7.53 & 96.25 $\pm$ 0.20 & 18.63 $\pm$ 7.95 & 98.20 $\pm$ 0.07 \\
& SAGE & 37.83 $\pm$ 7.75 & 95.63 $\pm$ 0.27 & 47.84 $\pm$ 6.39 & 97.39 $\pm$ 0.15 & 22.74 $\pm$ 5.47 & 98.87 $\pm$ 0.04 \\
& GAE & 29.98 $\pm$ 3.21 & 95.08 $\pm$ 0.33 & 63.33 $\pm$ 3.14 & 97.06 $\pm$ 0.22 & 16.67 $\pm$ 0.19 & 97.47 $\pm$ 0.08 \\
\midrule
\multirow{6}{*}{\textbf{GNN + Pairwise Info}} 

& SEAL & 26.69 $\pm$ 5.89 & 90.59 $\pm$ 0.75 & 39.36 $\pm$ 4.99 & 88.52 $\pm$ 1.40 & 38.06 $\pm$ 5.18 & 97.77 $\pm$ 0.40 \\
& BUDDY & 26.40 $\pm$ 4.40 & 95.06 $\pm$ 1.67 & 59.48 $\pm$ 8.96 & 96.72 $\pm$ 0.26 & 23.98 $\pm$ 5.11 & 98.2 $\pm$ 0.05 \\
& Neo-GNN & 22.65 $\pm$ 2.60 & 93.73 $\pm$ 0.36 & 53.97 $\pm$ 5.88 & 94.89 $\pm$ 0.60 & 31.45 $\pm$ 3.17 & 98.71 $\pm$ 0.05 \\
& NCN & 32.93 $\pm$ 3.83 & 96.76 $\pm$ 0.18 & 54.97 $\pm$ 6.03 & 97.04 $\pm$ 0.26 & 35.65 $\pm$ 4.60 & 98.98 $\pm$ 0.04 \\
& NCNC & 29.01 $\pm$ 3.83 & \colorbox{blue!20}{\textbf{96.90 $\pm$ 0.28}} & 64.03 $\pm$ 3.67 & 97.65 $\pm$ 0.30 & 25.70 $\pm$ 4.48 & 99.14 $\pm$ 0.03 \\
& NBFNet & 37.69 $\pm$ 3.97 & 92.85 $\pm$ 0.17 & 38.17 $\pm$ 3.06 & 91.06 $\pm$ 0.15 & \colorbox{blue!20}{\textbf{44.73 $\pm$ 2.12}} & 98.34 $\pm$ 0.02 \\
& PEG & 22.76 $\pm$ 1.84 & 94.46 $\pm$ 0.34 & 56.12 $\pm$ 6.62 & 96.15 $\pm$ 0.41 & 21.05 $\pm$ 2.85 & 96.97 $\pm$ 0.39 \\
\midrule
& GAGA(Ours) &  \colorbox{blue!20}{\textbf{46.22 $\pm$ 2.13}} & 96.78 $\pm$ 0.02 & \colorbox{blue!20}{\textbf{64.83 $\pm$ 3.11}} & \colorbox{blue!20}{\textbf{98.13 $\pm$ 0.11}} & 44.31 $\pm$ 2.12 & \colorbox{blue!20}{\textbf{99.24 $\pm$ 0.01}} \\
\bottomrule
\end{tabular}}
\end{table*}

\noindent  \textbf{Impact of the Prototype Size \(k_p\).} We used prototype embeddings to reduce the computational complexity during the downstream finetuning tasks. Here we will consider the effect of the prototype dimension. 
In Table \ref{Tab:diffk},  we can see while increasing \(k_p\) generally leads to higher computational costs, test accuracy remained relatively stable across different values. For instance, at \(k_p=10\), the model achieved a test accuracy of 76.42\% with the lowest time (201.01 seconds) and memory usage (5.07 GB). As \(k_p\) increased to 1280, the accuracy slightly improved to 76.71\%, but this came at the expense of significantly higher time (641.31 seconds) and memory (17.14 GB) requirements. Thus, the prototype projection is highly efficient since $k_p=10$ already brings a good performance.

\begin{table}[htbp]
\centering
\caption{Experimental results for different prototype dimension $k_p$ on ogbn-arxiv. Time and Memory are for the fine-tuning stage.}
\label{Tab:diffk}
\resizebox{0.55\linewidth}{!}{
\begin{tabular}{cccc}
\toprule
\bm{$k_p$} & \textbf{Memory (GB)} & \textbf{Time (s)} & \textbf{Test Accuracy(\%)} \\
\midrule
10    & 5.07  & 201.01 $\pm$ 8.61 & 76.42 $\pm$ 0.07  \\
40    & 5.24  & 212.43 $\pm$ 0.21 & 76.65 $\pm$ 0.01\\
80    & 5.47  & 223.03 $\pm$ 2.73 & 76.56 $\pm$ 0.09\\
160   & 5.92  & 242.37 $\pm$ 5.41 & 76.45 $\pm$ 0.05\\
320   & 7.44  & 291.87 $\pm$ 8.80 & 76.40 $\pm$ 0.03\\
640   & 10.67 & 413.12 $\pm$ 2.21 & 76.57 $\pm$ 0.10\\
1280  & 17.14 & 641.31 $\pm$ 6.17 & 76.71 $\pm$ 0.02\\
1693  & 21.31 & 782.28 $\pm$ 2.38 & 76.66 $\pm$ 0.06\\
\bottomrule
\end{tabular}
}
\end{table}

\vspace{-7pt}
\subsection{Ablation Study}

\noindent  \textbf{Impact of the GNN Backbone on Performance.}  Since our framework is plug-and-play and does not alter the structure of the GNN model, it can be used with any GNN backbone. We explored the impact of different GNN backbones on classification accuracy.  We tested the effects on several different backbones, including MLP, GCN, SAGE, GAT, and RevGAT in Table~\ref{Tab:backbones}.  

While using various GNNs as backbones did affect the final classification accuracy of the model, the impact was not significant. The only exception is MLP; its accuracy is lower, maybe because it is too simple to capture the relationship between the nodes. This demonstrates the robustness of our model.

\noindent  \textbf{Impact of Number of Selected Nodes.} 
We study the effect of different proportions of selected nodes on performance. As shown in Table \ref{Tab:ratio}, while increasing the number of selected nodes can slightly improve classification accuracy,  the model can already achieve strong performance ($0.7571$) even with just 0.1\% annotated data. This efficiency is due to redundancy in the nodes' textual information. Specifically, in ogbn-arxiv, abstracts and titles of papers in the same category are highly similar, enabling effective learning from limited data.  The use of high-information-density data allows the model to learn quality representations with fewer examples.

\vspace{-7pt}
\section{Conclusions}
\vspace{-7pt}
We proposed GAGA, a lightweight framework for node classification and link prediction in Text-attributed Graphs that minimizes annotation effort by focusing on 1\% of representative nodes and edges. Through structural alignment and efficient annotation, GAGA achieves state-of-the-art accuracy with significantly reduced annotation cost.

\newpage 

\bibliographystyle{plain}
\bibliography{nips}


\appendix

\newpage 
\section{Additional Experimental Details}
\label{App:add_exp}
\subsection{Datasets} 
\label{App: datasets}
\begin{table}[ht]
    \centering
    \caption{Statistics of TAGs used in this work.}
    \begin{tabular}{lccc}
        \toprule
        \textbf{Dataset} & \textbf{Nodes} & \textbf{Edges} & \textbf{Classes} \\
        \midrule
        Cora & 2,708 & 5,429 & 7 \\
        PubMed & 19,717 & 44,338 & 3 \\
        ogbn-arxiv & 169,343 & 1,166,243 & 40 \\
        CiteSeer & 3,312 & 4,732 & 6 \\
        ogbn-products & 54,025 &  74,420 & 47 \\
        tape-arxiv23 & 46,198 & 78,543 & 40 \\
        \bottomrule
    \end{tabular}
    \label{tab:dataset_summary}
\end{table} 
The details of the datasets are as follows:

\noindent \textbf{Cora.} The Cora dataset \citep{mccallum2000automating} consists of 2,708 scientific publications categorized into seven distinct classes: case-based, genetic algorithms, neural networks, probabilistic methods, reinforcement learning, rule learning, and theory. This dataset includes a citation network comprising 5,429 links. The selection process ensured that each paper in the final corpus either cites or is cited by at least one other paper.

\noindent \textbf{PubMed.} The PubMed dataset \citep{sen2008collective} comprises 19,717 scientific publications from the PubMed database, all related to diabetes. These publications are categorized into three distinct classes: Experimental Induced Diabetes, Type 1 Diabetes, and Type 2 Diabetes. The dataset also includes a citation network with 44,338 links.

\noindent \textbf{ogbn-arxiv.} The ogbn-arxiv dataset \citep{hu2020open} is a directed graph representing the citation network among all computer science papers on arXiv, as indexed by MAG (Wang et al., 2020). In this dataset, each node corresponds to an arXiv paper, and each directed edge signifies a citation from one paper to another. The primary task is to predict the 40 subject areas of the arXiv computer science papers, such as cs.AI, cs.LG, and cs.OS, which are manually assigned by the authors and arXiv moderators.

\noindent \textbf{Citeseer.} The CiteSeer dataset \citep{10.1145/276675.276685} consists of 3312 scientific publications classified into one of six classes. The citation network consists of 4732 links. Each publication in the dataset is described by a 0/1-valued word vector indicating the absence/presence of the corresponding word from the dictionary. The dictionary consists of 3703 unique words.

\noindent \textbf{ogbn-products} The ogbn-products \cite{hu2020open} dataset is an Amazon product co-purchasing network with products as nodes and edges indicating co-purchases. The task is to predict the product category in a multi-class setup using 47 top-level categories as labels.

\noindent \textbf{tape-arxiv23} The tape-arxiv23 \cite{he2023harnessing} dataset is a directed graph of citation networks among computer science arXiv papers from 2023 onwards. Each node represents a paper, and the edges show citation links. The task is to predict the 40 subject areas of these papers, such as cs.AI, cs.LG, and cs.OS, based on author and moderator labels.

\subsection{Baselines}
\label{App:baselines}
\subsubsection{Node Classification}
Details about the baseline models for node classification are as follows:

\noindent \textbf{MLP.} As a baseline model, MLP does not consider graph structure information and relies solely on node features for classification. Its performance provides a reference point for assessing the improvements offered by other models.

\noindent \textbf{GCN.} GCN~\citep{kipf2016semi} is a classical graph neural network model that effectively captures local neighborhood information through graph convolution operations. We chose GCN due to its strong performance and widespread application across many graph datasets.

\noindent \textbf{GraphSAGE.} GraphSAGE~\citep{sun2021scalable} generates node embeddings by sampling and aggregating information from neighboring nodes, making it suitable for handling large-scale graph data. We used GraphSAGE to evaluate its generalization capability across different datasets. 

\noindent \textbf{RevGAT.} RevGAT~\citep{li2021training} combines graph attention mechanisms with reversible network structures, reducing memory consumption while maintaining performance. We selected RevGAT to explore the effectiveness of attention mechanisms on graph data.

\noindent \textbf{InstructGLM.} InstructGLM \cite{ye2023natural}  integrates the strengths of graph neural networks and language models, enhancing generalization through instruction learning. We here use InstructGLM using Llama-7b \citep{touvron2023llama} as the backbone.

\noindent \textbf{Graphormer.} Graphormer~\citep{ying2021transformers} leverages the Transformer architecture to process graph data, capturing global dependencies. We selected Graphormer to test the performance of Transformer-based approaches in graph neural networks.

\noindent \textbf{TAPE.} The TAPE model~\citep{he2023harnessing}  leverages LLMs to capture textual information as features, which can be used to boost GNN performance on downstream tasks. A key innovation is its use of explanations as features: TAPE prompts an LLM to perform zero-shot classification, requests textual explanations for its decision-making process, and designs an LLM-to-LM interpreter to translate these explanations into informative features for downstream GNNs. For fairness, we use TAPE with the backbone of the GCN model, which is the same as our model.

\noindent \textbf{GLEM.} GLEM \citep{zhao2022learning} combines graph models with LLMs, like DeBERTa [12], in a variational EM framework, alternating between updating the LLM and GNN in the E-step and M-step to enhance downstream task performance.

\noindent \textbf{OneForAll.} OneForAll \citep{liu2023one} represents all nodes and edges as human-readable texts and encodes them from various domains into a unified space using LLMs. The framework then adapts to different tasks by adding task-specific prompts into the input graph.

\noindent \textbf{ENGINE.} ENGINE \citep{zhu2024efficient} introduces a lightweight, tunable G-Ladder module to each LLM layer, using a message-passing mechanism to incorporate structural information. This allows token-level outputs from each LLM layer to pass through the G-Ladder, enhancing node representations for node classification.

\noindent \textbf{GIANT.} GIANT \citep{chien2021node} utilizes XR-Transformers \citep{zhang2021fast} for neighborhood prediction, producing an LLM that generates more effective feature vectors for node classification than bag-of-words and vanilla BERT embeddings.

\noindent \textbf{SimTEG.} SimTeG \citep{duan2023simteg} fine-tunes a pre-trained LM with parameter-efficient methods for a task like node classification, then uses the LM’s last hidden states as node embeddings for GNN training, significantly boosting performance on graph benchmarks.

\subsubsection{Link Prediction}
For the link prediction tasks, we use the baseline models from HeaRT \citep{li2023evaluating,mao2023revisiting}. 

\noindent  \textbf{Heuristic methods}: Common Neighbor (CN) \citep{newman2001clustering}, Adamic Adar (AA) \citep{adamic2003friends}, and Resource Allocation (RA) \citep{zhou2009predicting} use common neighbors, while Shortest Path \citep{liben2003link} and Katz \citep{katz1953new} rely on path information to score link existence. 

\noindent  \textbf{Embedding methods}: Matrix Factorization (MF) \citep{menon2011link}, Multilayer Perceptron (MLP), and Node2Vec \citep{grover2016node2vec} learn low-dimensional node embeddings to predict link likelihood. 

\noindent  \textbf{GNN methods}: Graph Convolutional Network (GCN) \citep{kipf2016semi}, Graph Attention Network (GAT) \citep{veličković2018graph}, GraphSAGE (SAGE) \citep{sun2021scalable}, and Graph Autoencoder (GAE) \citep{kipf2016variational} integrate multi-hop graph structures via message passing. 

\noindent  \textbf{GNN + Pairwise Information}: SEAL \citep{zhang2021labeling}, BUDDY \citep{chamberlain2022graph}, and Neural Bellman-Ford Network (NBFNet) \citep{zhu2021neural} use subgraph features; Neo-GNN \citep{yun2021neo}, Neighborhood Contrastive Network (NCN) \citep{wang2023neural}, and NCNC \citep{wang2023neural} utilize common neighbor information, while Position Encoding Graph Neural Network (PEG) \citep{wang2022equivariant} employs positional encoding.

\subsection{Experiment Setting Details}
\subsubsection{Baseline settings}
For details on the settings of each baseline, we use their default hyper-parameters and settings, please refer to their respective GitHub repositories:
\begin{itemize}
    \item \textbf{MLP, GCN, SAGE}: \url{https://github.com/snap-stanford/ogb/tree/master/examples/nodeproppred/arxiv}
    \item \textbf{TAPE, RevGAT}: \url{https://github.com/XiaoxinHe/TAPE/tree/main}
    \item \textbf{Graphormer}: \url{https://github.com/Microsoft/Graphormer}
    \item \textbf{InstructGLM}: \url{https://github.com/agiresearch/InstructGLM}
    \item \textbf{GLEM}: \url{https://github.com/AndyJZhao/GLEM}
    \item \textbf{SimTEG}: \url{https://github.com/vermouthdky/SimTeG}
    \item \textbf{ENGINE}: \url{https://github.com/ZhuYun97/ENGINE}
    \item \textbf{GIANT}: \url{https://github.com/amzn/pecos/tree/mainline/examples/giant-xrt}
    \item \textbf{HeaRT}: \url{https://github.com/Juanhui28/HeaRT}
\end{itemize}

\subsubsection{Time and Cost Comparison Settings}
\label{app:time_money_settings}
In this Time and Cost Comparison experiment, \texttt{accuracy} refers to the node classification accuracy. {\texttt{money}} includes the computing costs, such as renting a V100 GPU instance on platforms like AWS (approximately \$3.06 per hour at the time of our experiments, though this rate may have changed), as well as the costs for OpenAI API token consumption. The \texttt{time} includes the total duration of the process, encompassing the LLM annotation phase (if applicable), the subsequent alignment process (if applicable) and fine-tuning process.

\subsection{Evaluation Metrics} \label{App: metrics}
We use MRR@10 and AUC for link prediction tasks.
\begin{equation*}
MRR = \frac{1}{N} \sum_{i=1}^{N} \frac{1}{\text{rank}_i}, 
\end{equation*}
where  $N$ is the number of positive samples and $\text{rank}_i$ is the rank of the $i$-th positive sample. 
\begin{equation*}
AUC = \frac{1}{|\mathcal{D}^0| \times |\mathcal{D}^1|} \sum_{i \in \mathcal{D}^0} \sum_{j \in \mathcal{D}^1} \mathbb{I}(\text{score}_i > \text{score}_j),  
\end{equation*}
where $\mathcal{D}^0$ is the set of positive samples, $\mathcal{D}^1$ is the set of negative samples, and $\mathbb{I}$ is the indicator function.

\section{Additional Experiments}\label{sec:appen_results}

\noindent  \textbf{Impact of Text Encoder.} In this section, we aim to investigate the impact of different language models on the performance of GAGA. We replace the language model with various alternatives and compare their performance on the ogbn-arxiv dataset. In GAGA, we align annotations and graphs, both of which are represented by features extracted from a language model. Therefore, the selection of an appropriate language model significantly impacts the final performance of the model. 

We also compared the effects of different language models on the overall classification accuracy. The results in Table \ref{Tab:lmbackbone}
showed that using larger models may not improve classification accuracy. This may be due to the high dimensionality of embeddings in larger models, which may hinder the learning process of the GNN. Furthermore, due to the limited data utilized during the alignment process, it is possible that larger language models were not adequately aligned, contributing to the suboptimal performance observed.
\begin{table*}[ht]
\centering
\caption{Ablation study on the effects of different LLMs for annotation. }
\label{Tab:lmbackbone}
\resizebox{0.85\linewidth}{!}{
\begin{tabularx}{\textwidth}{X X X X X X}
\toprule
Backbone & all-MiniLM-L6-v2 & MiniLM-L12-H384-uncased & gte-Qwen2-1.5B-instruct & gte-Qwen2-7B-instruct & LLM2Vec-Meta-Llama-3-supervised \\
\midrule
Model Size & 23M & 33M & 1.5B & 7B & 8B \\
\midrule
Test & 0.7620 & 0.7681 & 0.7253 & 0.7515 & 0.7519 \\
Valid & 0.7701 & 0.7722 & 0.7404 & 0.7531 & 0.7556 \\
\bottomrule
\end{tabularx}}
\end{table*}

\begin{figure}[htbp]
  \centering
  \includegraphics[width=0.5\textwidth]{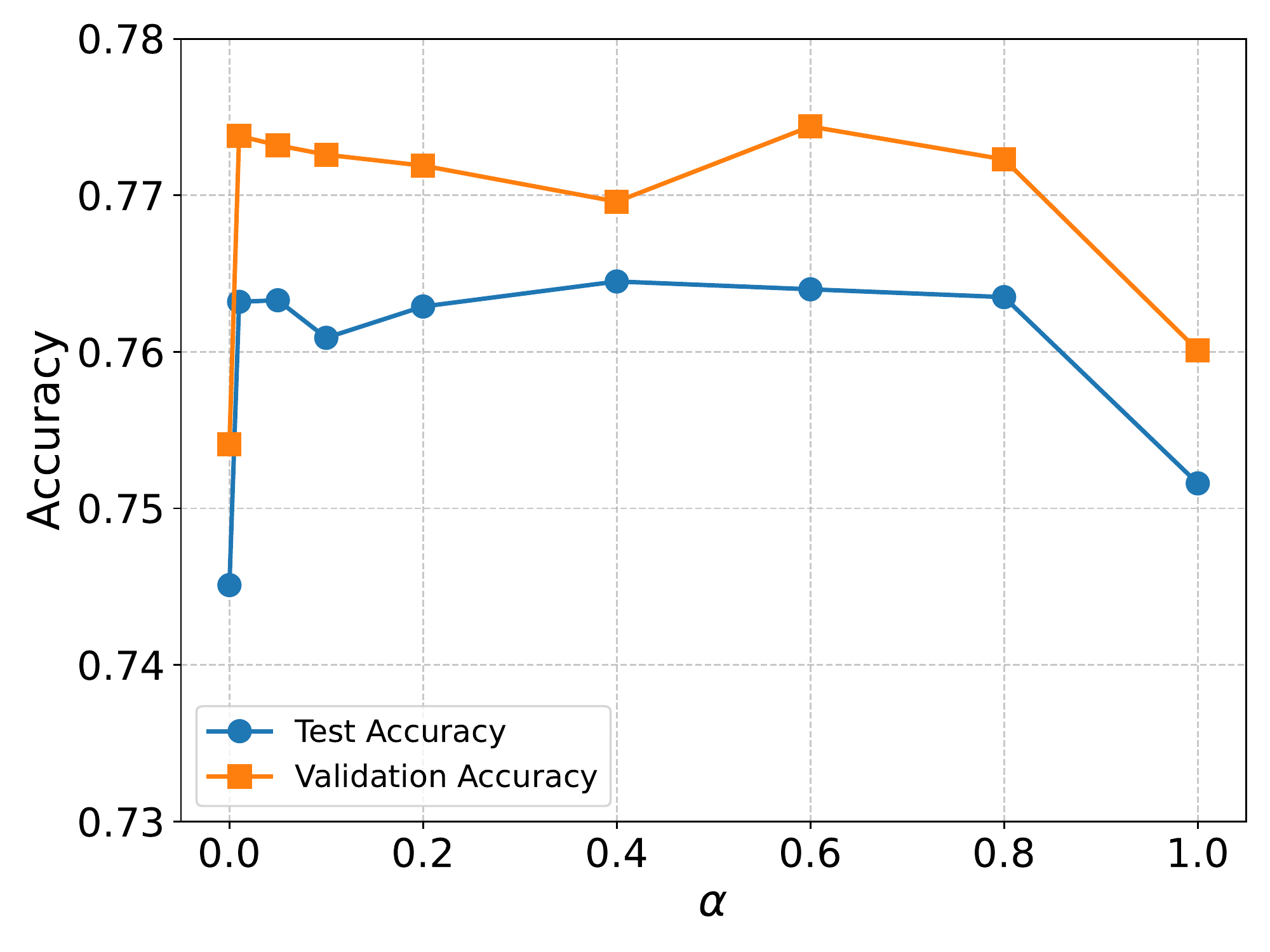}
  \caption{Impact of \(\alpha\) on Valid and Test Accuracy on ogbn-arxiv Dataset.}
  \label{fig:alpha}
\end{figure}

\noindent  \textbf{Impact of \(\alpha\).} In Figure \ref{fig:alpha}, we studied the effect of the hyperparameter \( \alpha \)  in equation (\ref{eq:3}), which adjusts the model’s focus between the alignment of sub annotation graph and sub textual graph, and the alignment of sub textual graph and prototype projections. As shown in the figure, when \( \alpha \) is set to 0, the model disregards the prototypes and focuses on the original contrastive
loss, making generalization ability worse as the model will learn too much redundant information.  When \( \alpha \) is set to 1, it focuses more on prototype alignment and ignores the original sub annotation graphs, making the utility
worse as the project may lose too much information. Both extreme cases result in a decline in model performance. However, when \( \alpha \) is between 0 and 1, the model’s performance remains relatively stable.

\noindent \textbf{Impact of LLMs.} We conducted experiments with meta-llama/Meta-Llama-3.1-8B-Instruct (Llama 3.1 8B) and Qwen/Qwen2.5-7B-Instruct (Qwen 2.5 7B). The validation and test accuracies for each model on ogbn-arxiv dataset are summarized in Table~\ref{tab:results}.

\begin{table*}[h!]
\centering
\caption{Validation and Test Accuracies of Different LLMs (in \%)}
\label{tab:results}
\begin{tabular}{@{}lcc@{}}
\toprule
\textbf{LLM}    & \textbf{Validation} & \textbf{Test} \\ \midrule
Llama 3.1 8B    & $77.02 \pm 0.09$             & $76.06 \pm 0.11$       \\
GPT3.5-turbo          & $77.09 \pm 0.05$             & $76.01 \pm 0.04$       \\
Qwen 2.5 7B     & $76.98 \pm 0.14$             & $76.06 \pm 0.09$       \\
\bottomrule
\end{tabular}
\end{table*}

The experimental results indicate that our method demonstrates strong robustness in selecting large language models (LLMs). The validation and test accuracies across different models (Llama 3.1 8B, OpenAI, and Qwen 2.5 7B) are very similar, ranging from approximately $0.7698$ to $0.7709$. This consistency suggests that our approach performs reliably regardless of the specific model chosen.

Moreover, the low standard deviations in the accuracy scores indicate that the models exhibit minimal variability in performance across multiple runs. This stability is a key indicator of robustness, showing that our method can maintain high performance even when faced with different input data or model selections.

Overall, the results provide confidence that our method is effective and adaptable, ensuring reliable outcomes across various LLMs.

\noindent  \textbf{Impact of Different Feature Combination Methods.}
During our downstream task fine-tuning, the rationale behind applying an expensive attention operator to combine features remained unclear. A relevant question was raised: what if we simply concatenate prototype features and annotate graph features instead?

To address this, we conducted additional experiments comparing different feature combination approaches---cross-attention, concatenation, and averaging---on the arxiv-ogbn dataset. Our results from five runs demonstrate that cross-attention consistently outperforms other methods:

\begin{table}[h!]
\centering
\small
\begin{tabularx}{0.8\textwidth}{>{\centering\arraybackslash}X 
                                 >{\centering\arraybackslash}X 
                                 >{\centering\arraybackslash}X} 
\toprule
Method & Evaluation Accuracy (\%) & Test Accuracy (\%)\\
\midrule
Cross-Attention & 77.82 $\pm$ 0.04 & 76.27 $\pm$ 0.12 \\
Concatenation & 72.91 $\pm$ 0.11 & 71.08 $\pm$ 0.06 \\
Average & 72.01 $\pm$ 0.11 & 71.98 $\pm$ 0.06 \\
Baseline (GCN) & 71.93 $\pm$ 0.05 & 71.82 $\pm$ 0.20 \\
\bottomrule
\end{tabularx}
\caption{Comparison of feature combination approaches on the arxiv-ogbn dataset.}
\label{tab:cross-attn-results}
\end{table}

These results indicate that cross-attention is the most effective feature combination approach in this setting.

\section{Visualization of  Annotation Prototypes}\label{sec:anno}
In our study, we propose the use of annotation prototypes based on the observation that there is semantic clustering among concepts. When aligning node representations, the alignment can be performed in the representation space of specific concepts and the space obtained after semantic aggregation through prototypes.

To evaluate the effectiveness of semantic aggregation, we employed GPT sampling for assessment. Specifically, on the OGBN-Arxiv dataset, we randomly selected 1,000 nodes and calculated the distance between these nodes and the prototypes. For each prototype, we identified the 10 nodes closest to it, considering these nodes as a cluster. We then used GPT-4 to label these clusters. The prompt used for this purpose was:

\begin{tcolorbox}[colback=gray!5!white, colframe=gray!75!black, title=Prompt]
{node\_texts}

List three key words shared by the papers above.
\end{tcolorbox}

\begin{figure*}[htbp]
    \centering
    \subfigure[]{
        \includegraphics[width=0.48\textwidth]{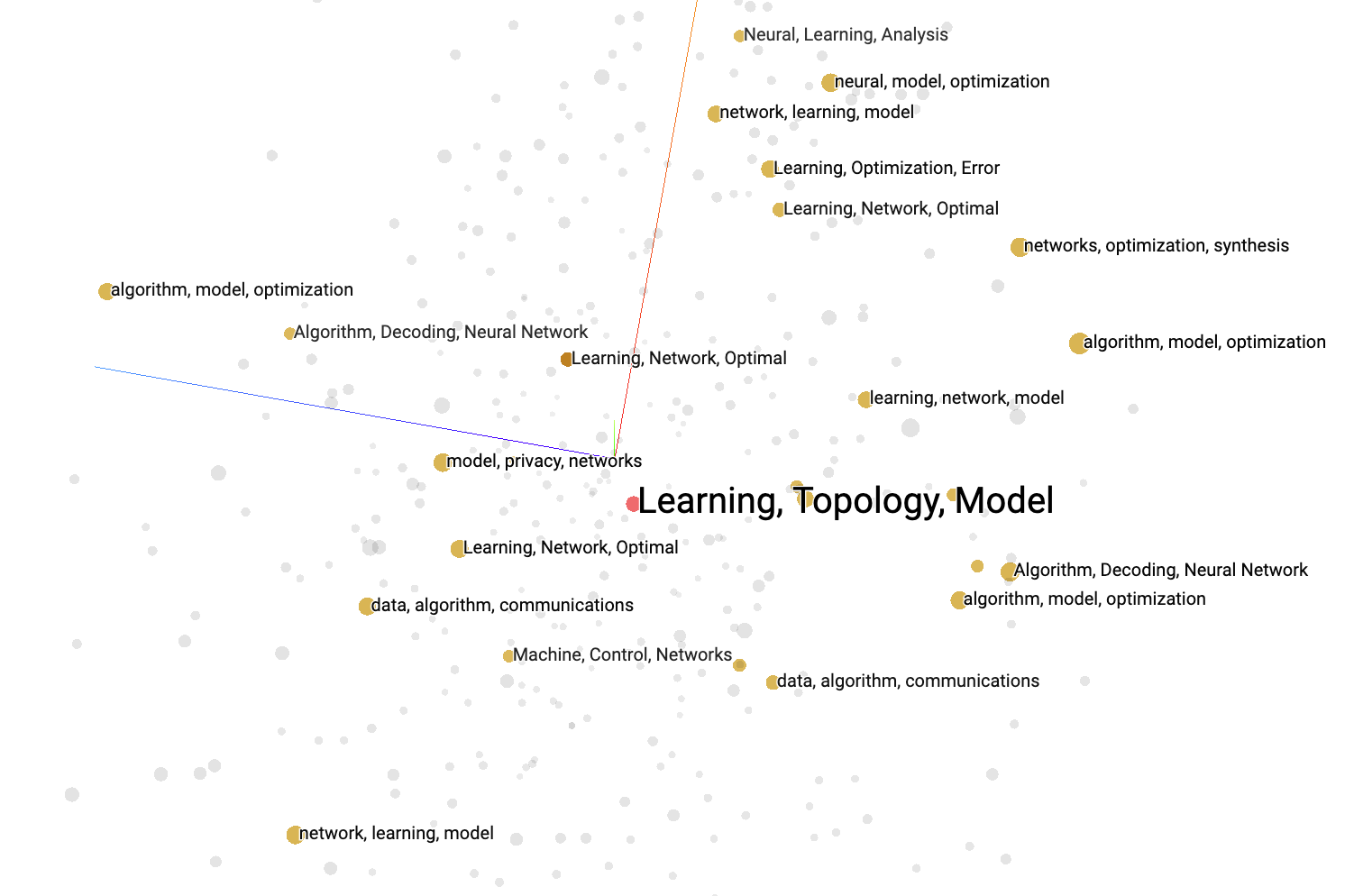}
    }
    \hfill
    \subfigure[]{
        \includegraphics[width=0.48\textwidth]{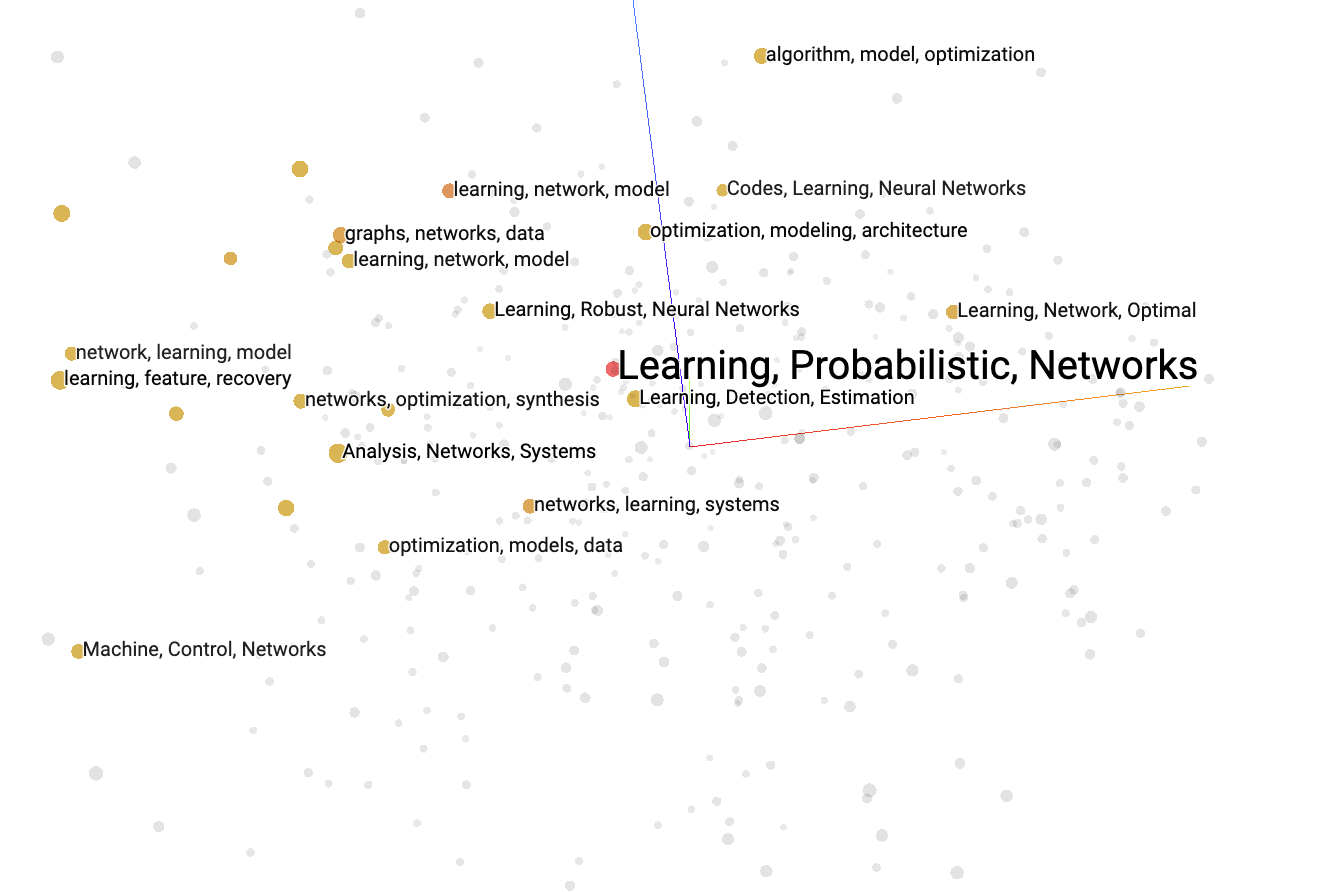}
    }
    \caption{Visualization of semantic aggregation using annotation prototypes.}
    \label{fig:prot_vis}
\end{figure*}

As shown in figure \ref{fig:prot_vis}, after obtaining the semantic labels, we applied Principal Component Analysis (PCA) for dimensionality reduction to a 3D space. The results demonstrated that nodes with similar semantic labels (prototypes) are also positioned close to each other in the semantic space, confirming the effectiveness of our semantic aggregation approach.

\begin{figure*}[htbp]
    \centering
    \subfigure[]{
        \includegraphics[width=0.45\textwidth]{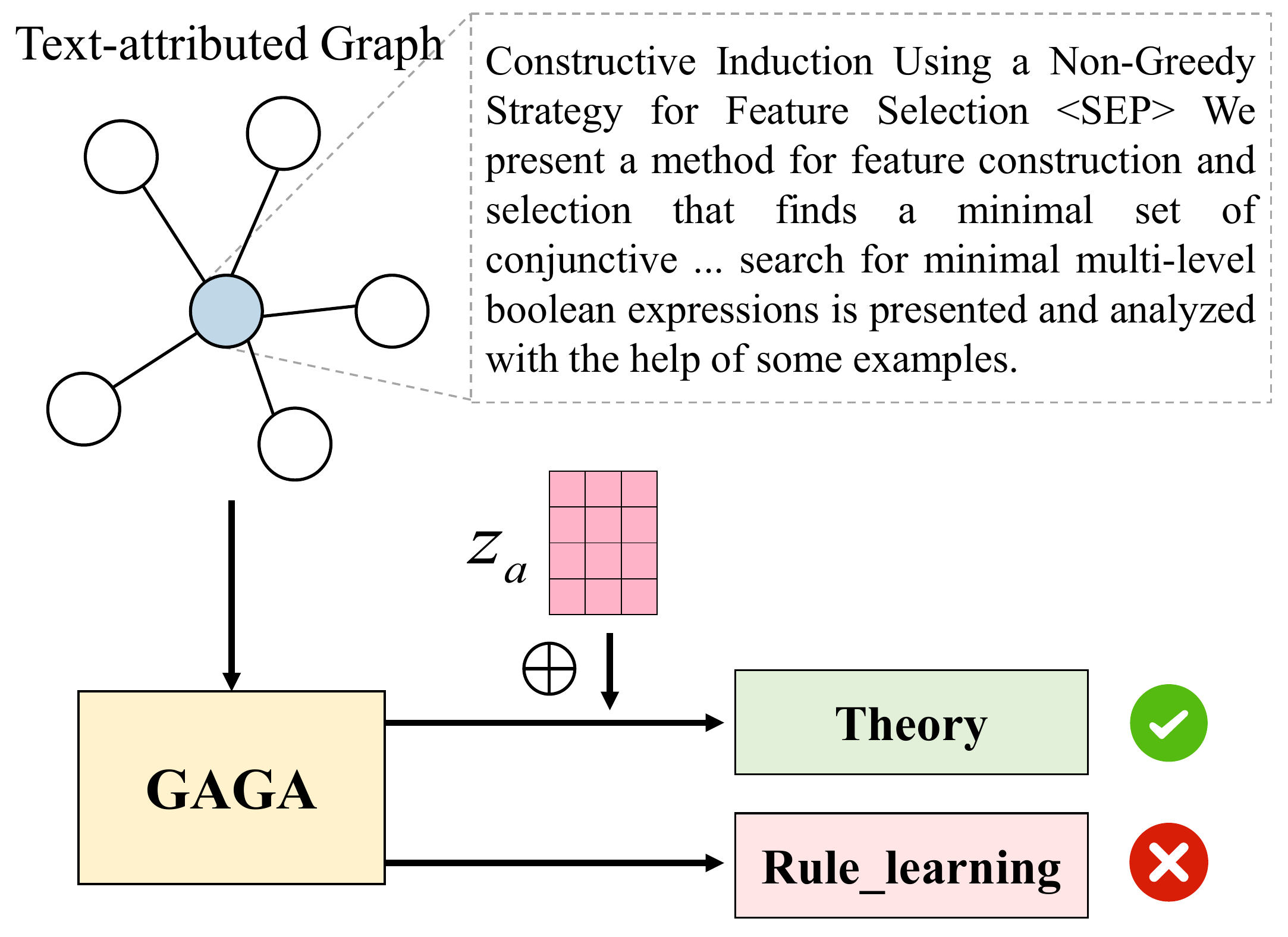}
        \label{fig:subfig1}
    }
    \hfill
    \subfigure[]{
        \includegraphics[width=0.45\textwidth]{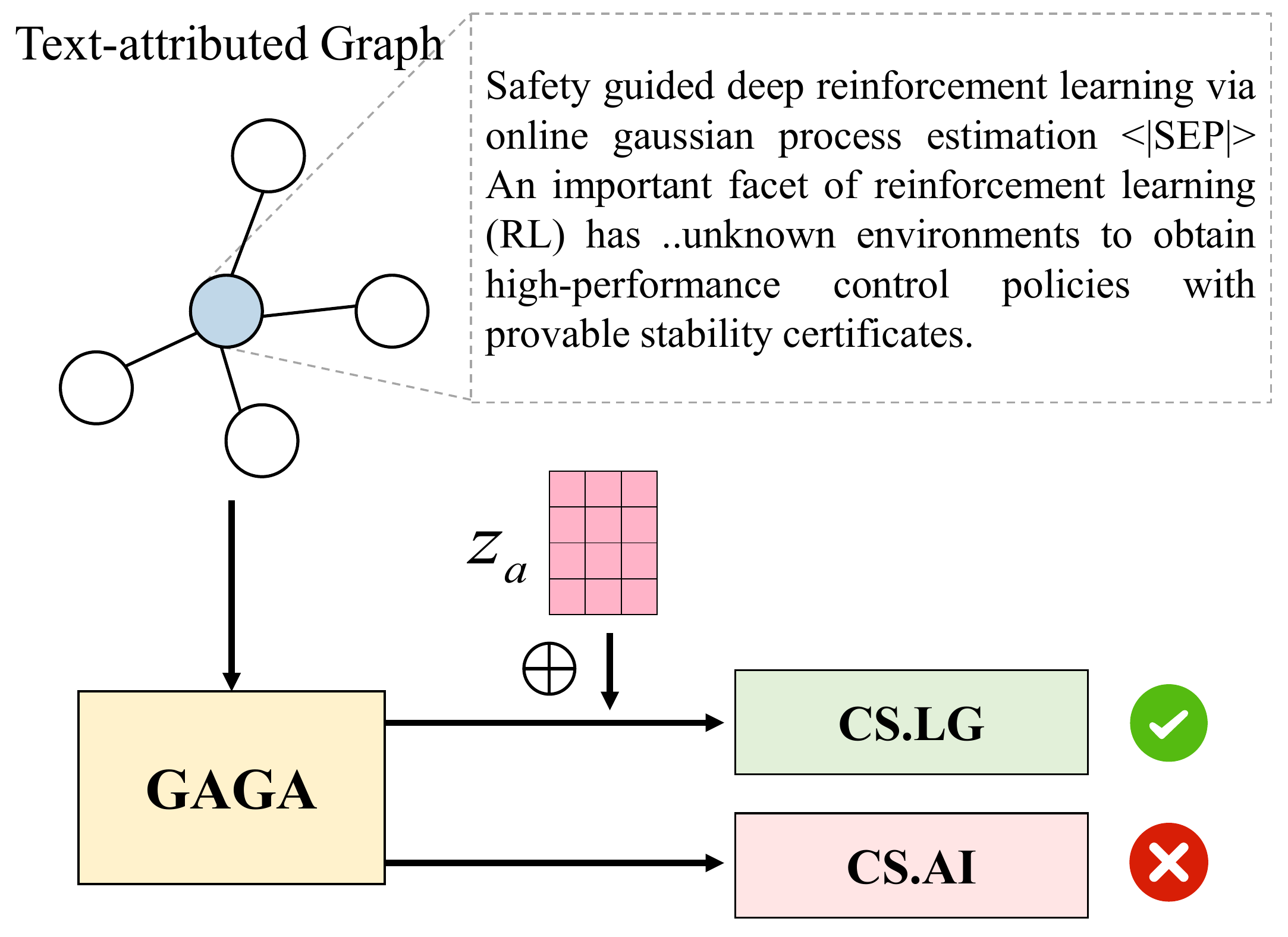}
        \label{fig:subfig2}
    }
    
    \vspace{1em}
    
    \subfigure[]{
        \includegraphics[width=0.45\textwidth]{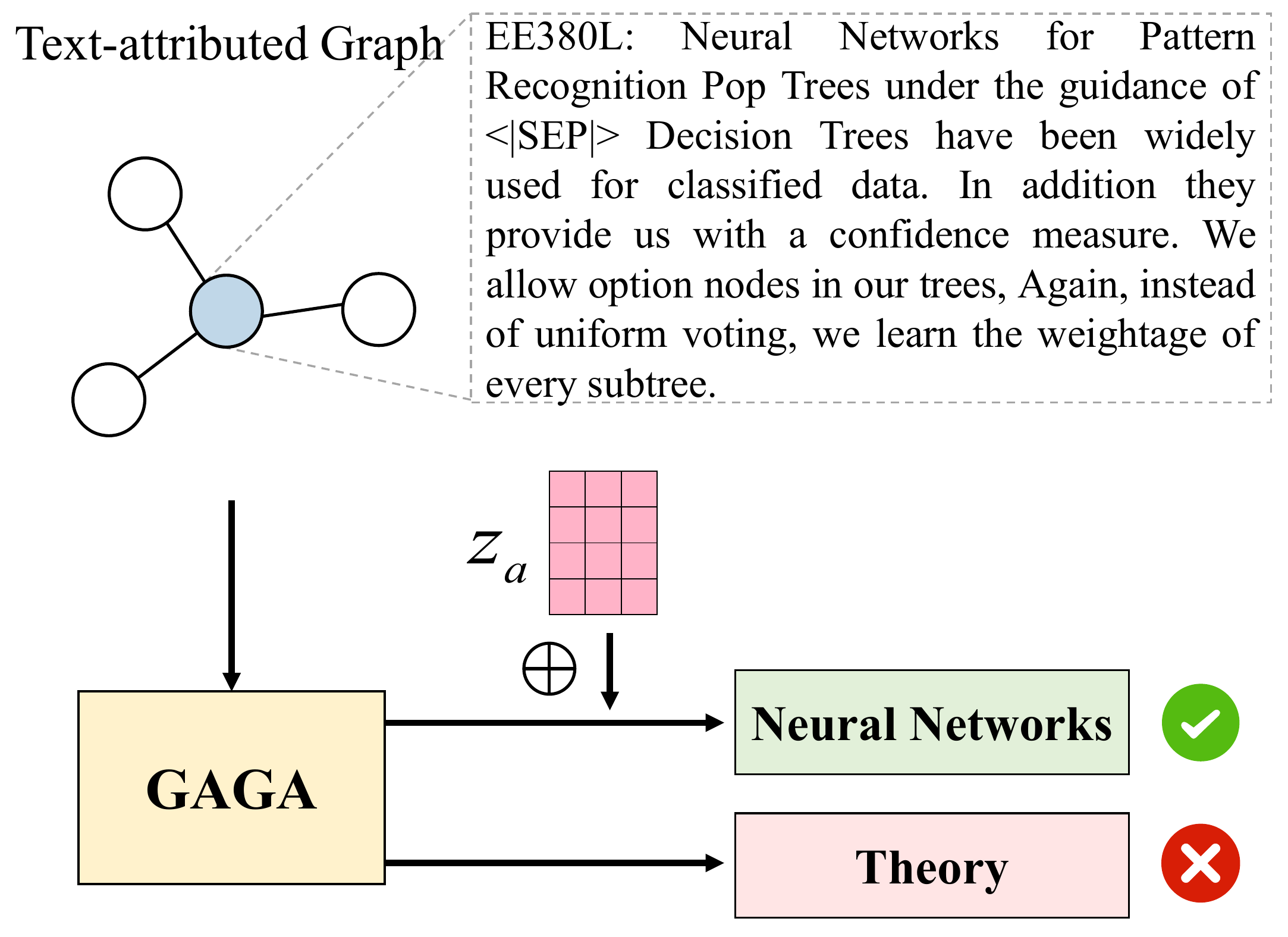}
        \label{fig:subfig3}
    }
    \hfill
    \subfigure[]{
        \includegraphics[width=0.45\textwidth]{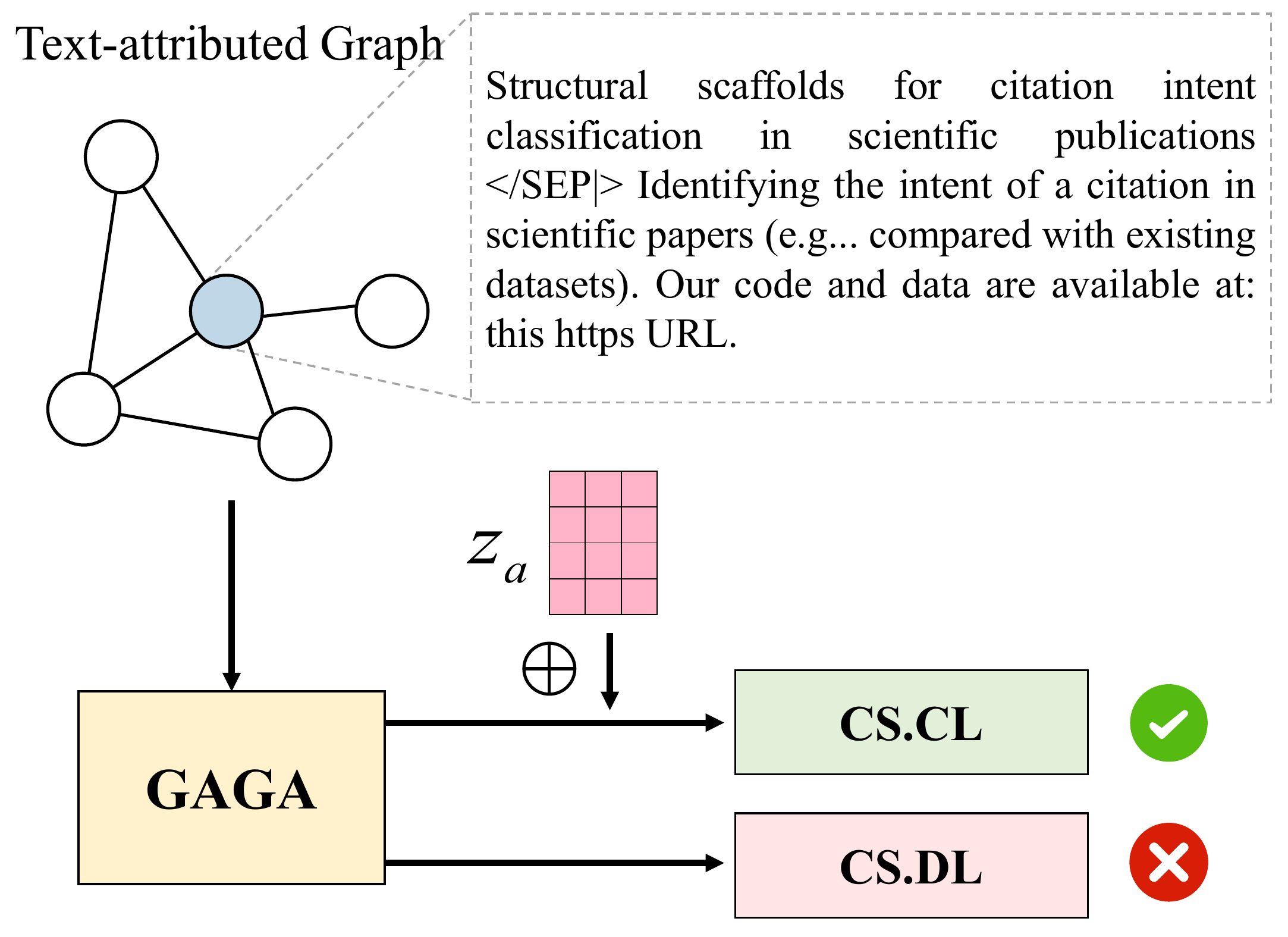}
        \label{fig:subfig4}
    }
    
    \caption{Visualization of the effect of annotation prototype projection.}
    \label{fig:all}
\end{figure*}

We also present several case studies, as shown in Figure \ref{fig:all}, to illustrate how the annotation prototype enhances the model's performance. When the input textual information is processed by GAGA, directly predicting the category of the node can sometimes result in inaccurate predictions. However, by incorporating prototype information and integrating it into the model's prediction through a cross-attention mechanism, the model can effectively adjust and refine its predictions. This is because the integration allows the model to leverage additional contextual knowledge from the prototypes, thereby enhancing its ability to correct initial errors and produce more accurate final predictions.

\section{Prompts for Generating Annotations}
In this section, we present our prompts used to generate annotations.
\label{App:prompts}
\begin{tcolorbox}[title=Prompt Template, colframe=black, colback=white, boxsep=1mm, left=1mm, right=1mm, top=0.5mm, bottom=0.5mm, width=\linewidth]
\{Node's textual information\}
[Question]: Which of the following category does this node belong to: \{possible categories\}. 

Give 5 likely categories as a comma-separated list ordered from most to least likely. \\ 
List the most important concepts in the paper.
And you should tell me what knowledge is needed to understand the concepts. After all, you should provide your reasoning. 

Your response:
\end{tcolorbox}
\begin{tcolorbox}[title=Prompt for node classification on ogbn-arxiv.]
Abstract: \{abstract\}
Title: \{title\}
Question: Which arXiv CS subcategory does this paper belong to? 
Give 5 likely arXiv CS sub-categories as a comma-separated list ordered from most to least likely, 
in the form “cs.XX”, list the most important concepts in the paper. 
And you should tell me what knowledge is needed to understand the concepts. 
After all, you should provide your reasoning.
Your response:
\end{tcolorbox}

\begin{tcolorbox}[title=Prompt for node classification on Arxiv-2023., breakable]
Abstract: \{abstract text\}
Title: \{title text\}
Question: Which arXiv CS subcategory does this paper belong to? Give 5 likely arXiv CS sub-categories from most to least likely, in the form “cs.XX”, List the most important concepts in the paper. And you should tell me what knowledge is needed to understand the concepts. After all, you should provide your reasoning.
Your response:
\end{tcolorbox}

\begin{tcolorbox}[title=Prompt for node classification on Cora., breakable]
Abstract: \{abstract text\}
Title: \{title text\}
Question: Which of the following sub-categories of AI does this paper belong to: Case Based, Genetic Algorithms, Neural Networks, Probabilistic Methods, Reinforcement Learning, Rule Learning, Theory? Explain how your prediction is present in the text. List the most important concepts in the paper,and tell me what knowledge is needed to understand the concepts. After all, you should provide your reasoning.Your response:
\end{tcolorbox}

\begin{tcolorbox}[title=Prompt for node classification on PubMed., breakable]

Abstract: \{abstract text\}

Title: \{title text\}

Question: Does the paper involve any cases of Type 1 diabetes, Type 2 diabetes, or Experimentally induced diabetes? Please give one or more answers of either Type 1 diabetes, Type 2 diabetes, or Experimentally induced diabetes; if multiple options apply, provide a comma-separated list ordered from most to least related, then for each choice you gave, give a detailed explanation with quotes from the text explaining why it is related to the chosen option. List the most important concepts in the paper. 

And you should tell me what knowledge is needed to understand the concepts. 

After all, you should provide your reasoning.
Your response:
\end{tcolorbox}

\begin{tcolorbox}[title=Prompt for link prediction., breakable]
Paper 1:
Title: \{title1\}
Abstract: \{abstract1\}
Paper 2:
Title: \{title2\}
Abstract: \{abstract2\}

Question: Why are these two papers related? list the most important concepts in the abstract. And you should tell me what knowledge is needed to understand the concepts. Based on the concepts, explain why they are related.

Your response:
\end{tcolorbox}

\begin{tcolorbox}[title=Prompt for node classification on ogbn-products., breakable]
Product description: \{product description\}

Question: Which of the following category does this product belong to: 1) Home \& Kitchen, 2) Health \& Personal Care, 3) Beauty, 4) Sports \& Outdoors, 5) Books, 6) Patio, Lawn \& Garden, 7) Toys \& Games, 8) CDs \& Vinyl, 9) Cell Phones \& Accessories, 10) Grocery \& Gourmet Food, 11)
Arts, Crafts \& Sewing, 12) Clothing, Shoes \& Jewelry, 13) Electronics, 14) Movies \& TV, 15) Software, 16) Video Games, 17) Automotive, 18) Pet Supplies, 19) Office Products, 20) Industrial \& Scientific, 21) Musical Instruments, 22) Tools \& Home Improvement, 23) Magazine Subscriptions, 24) Baby Products, 25) NAN, 26) Appliances, 27) Kitchen \& Dining, 28) Collectibles \& Fine Art, 29) All Beauty, 30)
Luxury Beauty, 31) Amazon Fashion, 32) Computers, 33) All Electronics, 34) Purchase Circles, 35) MP3 Players \& Accessories, 36) Gift Cards, 37) Office \& School Supplies, 38) Home Improvement, 39) Camera \& Photo, 40) GPS \& Navigation, 41) Digital Music, 42) Car Electronics, 43) Baby, 44) Kindle Store, 45) Kindle Apps, 46) Furniture \& Decor? Give 5 likely categories as a comma-separated list ordered from most to least likely, list the most important concepts. 

And you should tell me what knowledge is needed to understand the concepts. 

After all, you should provide your reasoning.
Your response:
\end{tcolorbox}

\section{Limitations} \label{sec_limitation}
Currently, to ensure efficiency and accuracy, GAGA requires alignment for each dataset individually and cannot be directly applied to new TAGs. In the future, developing a pre-trained model to learn a generalized enhanced node representation that can be directly applied to different graph datasets could represent a promising new direction.

\end{document}